\documentclass{article}

\PassOptionsToPackage{numbers, compress}{natbib}
\usepackage[final]{neurips_2020}

\usepackage[utf8]{inputenc} % allow utf-8 input
\usepackage[T1]{fontenc}    % use 8-bit T1 fonts
\usepackage{hyperref}       % hyperlinks
\usepackage{url}            % simple URL typesetting
\usepackage{booktabs}       % professional-quality tables
\usepackage{amsfonts}       % blackboard math symbols
\usepackage{nicefrac}       % compact symbols for 1/2, etc.
\usepackage{microtype}  % microtypography
\usepackage{amsmath}
\usepackage{float}
\usepackage{subcaption}
\usepackage[colorinlistoftodos]{todonotes}
\usepackage[acronym]{glossaries}

\def\x{{\mathbf x}}

\begin{document}
\title{Predicting Regional Locust Swarm Distribution with Recurrent Neural Networks}

% The \author macro works with any number of authors. There are two commands
% used to separate the names and addresses of multiple authors: \And and \AND.
%
% Using \And between authors leaves it to LaTeX to determine where to break the
% lines. Using \AND forces a line break at that point. So, if LaTeX puts 3 of 4
% authors names on the first line, and the last on the second line, try using
% \AND instead of \And before the third author name.
%\thanks{Use footnote for providing further information about author (webpage, alternative address)---\emph{not} for acknowledging funding agencies.}
%\thanks{The first three authors have equal contribution.} 
\author{%
  Hadia Mohmmed Osman Ahmed  Samil\thanks{Equal Contribution}
\\
  %African Master’s in Machine Intelligence (AMMI)\\
  African Institute for Mathematical Sciences (Rwanda)\\
  hsamil@aimsammi.org \\
   \And
   Annabelle Martin\footnotemark[1]\\
   University of Montreal, Mila \\
   %Montreal, Canada\\
   annabelle.martin@umontreal.ca
  % Affiliation \\
  % Address \\
  % \texttt{email} \\
   \AND
   Arnav Kumar Jain\footnotemark[1]\\
   Microsoft IDC\\
  % India\\
  arnavkj95@gmail.com \\
  % Affiliation \\
  % Address \\
  %\texttt{arnavkj95@gmail.com} \\
   \And
   Susan Amin\\
   McGill University, Mila\\
   %Montreal, Canada \\
   susan.amin@mail.mcgill.ca\\
  % Affiliation \\
  % Address \\
  % \texttt{email} \\
   \And
  Samira Ebrahimi Kahou     \\
  École de Technologie Supérieure, Mila, CIFAR\\
  %Montreal, Canada\\
  samira.ebrahimi-kahou@etsmtl.ca
  % Affiliation \\
  % Address \\
  % \texttt{email} \\
}

\glsdisablehyper

\newacronym{cnn}{CNN}{Convolutional Neural Network}
\newacronym{fao}{FAO}{Food and Agriculture Organization of the United Nations}
\newacronym{lstm}{LSTM}{Long Short-Term Memory}
\newacronym{rnn}{RNN}{Recurrent Neural Network}
\newacronym{maxent}{MaxEnt}{Maximum Entropy}
\newacronym{glm}{GLM}{Generalized Linear Model}

\maketitle
\begin{abstract}
  Locust infestation of some regions in the world, including Africa, Asia and Middle East has become a concerning issue that can affect the health and the lives of millions of people. In this respect, there have been attempts to resolve or reduce the severity of this problem via detection and monitoring of locust breeding areas using satellites and sensors, or the use of chemicals to prevent the formation of swarms. However, such methods have not been able to suppress the emergence and the collective behaviour of locusts. The ability to predict the location of the locust swarms prior to their formation, on the other hand, can help people get prepared and tackle the infestation issue more effectively. Here, we use machine learning to predict the location of locust swarms using the available data published by the Food and Agriculture Organization of the United Nations. The data includes the location of the observed swarms as well as environmental information, including soil moisture and the density of vegetation. The obtained results show that our proposed model can successfully, and with reasonable precision, predict the location of locust swarms, as well as their likely level of damage using a notion of density.
  
\end{abstract}

\section{Introduction}
Food insecurity is threatening more regions than ever around the world. Climate change, drought, flood and conflict are all drivers that put many countries at risk of acute food insecurity \cite{thompson2009addressing,connolly2016climate,schmidhuber2007global}. This situation is made worse in parts of Africa, Asia and the Middle East by desert locusts. A single adult desert locust can eat its own weight \cite{uvarov1928locusts}, and according to the \gls{fao}, swarms of flying locusts can eat the same amount of food in one day as 35,000 people and can travel hundreds of miles when looking for food. Thus, controlling the effects of swarms and protecting crops are necessary to ensure that affected regions don't face widespread starvation.

Several studies have focused on the factors that contribute to the formation and movement of the swarms \cite{despland2000small}. Some others have pushed further and tried to find ways to prevent the formation of the swarms through the use of chemicals \cite{Shi1343}. However, as preventing the breeding or suppressing the swarm behavior in an effective manner is not practical in the near future, we need to have a reliable model that can predict the locations that swarms will likely attack so that people can prepare themselves and confront the locusts effectively. In this regard, taking multidisciplinary approaches can be very advantageous. In the past decades, there have been theoretical studies on the motion of locusts using statistical tools as well as physical and biological data \cite{ariel2015locust}. With the improvement of computational resources over the past few years, some researchers have utilized machine learning algorithms to study the effects of land surface temperature \cite{gomez2019desert} and soil moisture in Western Africa \cite{gomez2018machine} and other countries \cite{gomez2020modelling} on the presence of solitarious locusts and their potential breeding areas, while some others \cite{Ye_2020} use \gls{cnn} models to detect the types of locusts based on field images. However, to the best of our knowledge, the prediction of locust \emph{swarm} location, spanning diverse areas on the earth, based on environmental conditions (\emph{e.g.} soil moisture and the availability of vegetation), has not been addressed in the previous studies. In particular, considering the large number (several millions) of locusts in each swarm, and their fast pace in terms of both their movement and vegetation destruction, the ability to predict their location is crucial in getting prepared in advance and confronting the locusts. 

In this study, we present our preliminary work on forecasting the location of locust swarms. In particular, we utilize the data related to the location of the observed locust swarms over the course of the past few decades as well as the ecology data, including the soil moisture and the vegetation density, to train a \gls{lstm} network, and subsequently use the trained model to predict the location of swarms. Our results show that our trained network can predict the location of swarm attacks as well as the intensity of locust infestation with reasonable precision.

\section{Related work}
There has been extensive field \cite{uvarov1977grasshoppers, uvarov1929phases, kennedy1939behaviour, kennedy1951migration, stower1963photographic} and laboratory \cite{ELLIS195991, ellis1963changes, gillett1973social, buhl2006disorder, ariel2014individual} research regarding locust behaviour and swarming. Ellis \cite{ELLIS195991} studies the conditions involved in the formation of swarms, as well as the behavior of gregarious and non-gregarious locusts. She confirms that solitary locusts take only a couple of hours to change behavior and get attracted to other locusts, leading to the formation of groups and potentially swarms. Locust collective movement has been modeled both under the assumption that locusts follow a usual animal collective movement model \cite{edelstein2001mathematical}, and under the assumption that locusts have their own behavior, interact and move following their own rules \cite{ariel2014individual}. In particular, Ariel \emph{et al.} \cite{ariel2014individual} take into account different factors that affect locusts' behaviour and movement, including temperature and the choice of locomotion (walking or flying), as well as the direction of movement. However, the environmental impact on the locusts roaming pattern has not been studied sufficiently yet \cite{ariel2015locust}.

Recently, machine learning has been successfully utilized in locust research. Ye \emph{et al.}~\cite{Ye_2020} use a \gls{cnn} model, in particular, a ResNet~\cite{he2016deep} with batch normalization~\cite{ioffe2015batch} and classify locusts based on their species with high accuracy. Other researchers have been able to predict and detect breeding areas based on climatic and soil data. For instance, Kimathi \emph{et al.} \cite{breedingeastafrica} utilize the data for temperature and rainfall, as well as soil data with a \gls{maxent} model to detect the most likely locust breeding spots in different countries across East Africa. In particular, they split their observations per country and train their model on three of them. They subsequently confirm the accuracy of their model by applying the trained model to the other countries in their data set. Another study by G{\'o}mez \emph{et al.} \cite{gomez2018machine} uses a \gls{glm} and a random forest model to study the relation between soil moisture and the presence of locust eggs and solitarious hoppers in Western Africa. In a later study, the authors further expand the range of the covered areas to 31 countries \cite{gomez2020modelling}.

\section{Methods}
In this section, we provide details regarding the sources of data we have used, as well as the network architecture and the method of learning we have adopted for this study.

\subsection{Data Source}
\label{DataSource}
We use the data sets gathered and published by \gls{fao} through the Locust Hub portal\footnote{https://locust-hub-hqfao.hub.arcgis.com}. The data sets contain information regarding the different phases and forms the desert locusts may take during their lifetime, as well as the dates and the locations of locust observations. They also involve the description of the observed locusts (\emph{e.g.} color, density, flying height, maturity, \emph{etc.}) and the ecological conditions. In particular, the data sets consist of different groups of locusts, namely, 1) \emph{hoppers}, defined as solitarious wingless nymphs; 2) \emph{bands}, which are wingless nymphs that form a band; 3) \emph{adults}, characterized as mature (adult) desert locusts that do not form a group or concentration; and 4) \emph{swarms}, which are mature desert locusts that form a swarm. The ecological information includes the condition and density of vegetation as well as the soil moisture. This information is available for the most affected regions and contains data related to most of North Africa, part of the Middle East and Asia. 

We combine the available data sets together, and add an additional column, named \emph{Locust type}. Each row in our data frame contains the information for a specific location identified with the location coordinates $[x,y]$. Due to the sparsity of the original data, we choose to grid the whole area into $100\times100$ regions and subsequently aggregate the information collected from each region for each month. In addition, we divide the data into training, validation, and test sets according to the year they were collected. The details of data split are shown in Table~\ref{table:dataStats}. In particular, the training set consists of the data taken from year $1985$ to mid-$2017$. The validation set is from mid-$2017$ to mid-$2019$ and test sets include the available data between mid-$2019$ and $2021$. Finally, we use the training data to train a \gls{rnn}, as discussed below.

\subsection{Model Architecture}
In this work, we leverage a \gls{rnn} for our task. Specifically, we train a \gls{lstm} \cite{hochreiter1997long} network, which consists of \gls{lstm} cells comprising of a cell state $c_t$ and a hidden state $h_t$. These states summarize information observed up to a certain timestep $t$. For updating these states, \gls{lstm} has three gates, named input($i_t$), output($o_t$), and forget($f_t$). They help with deciding the extent of the information flow to update the hidden state at the next time step, $h_{t+1}$. The recurrence of a \gls{lstm} cell can be described as:
\begin{equation}
\begin{split}
    	\mathbf i_t & = \sigma(\mathbf W_i\mathbf h_{t-1} + \mathbf U_i\x_t + \mathbf b_i) \\
		\mathbf f_t & = \sigma(\mathbf W_f\mathbf h_{t-1} + \mathbf U_f\x_t + \mathbf b_f) \\
		\mathbf o_t & = \sigma(\mathbf W_o\mathbf h_{t-1} + \mathbf U_o\x_t + \mathbf b_o) \\
		\mathbf{\tilde{c}_t} & = \tanh(\mathbf W_c\mathbf h_{t-1} + \mathbf U_c\x_t + \mathbf b_c) \\
		\mathbf c_t & =  \mathbf i_t \odot \mathbf{\tilde{c}_t} + \mathbf f_t \odot \mathbf c_{t-1}\\
        \mathbf h_t & = \mathbf o_t \odot \mathbf c_t,
\end{split}
\end{equation}
where $\odot$ denotes element-wise multiplication, and $\sigma$ is the sigmoidal non-linearity. $\mathbf W_x, \mathbf U_x, \mathbf A_x$ and $\mathbf b_x$, are the weight matrices for the previous hidden state, input, and bias, respectively. Further, we apply a single feed forward layer on the hidden state $h_t$:
\begin{equation}
    z_t = \mathbf W_ph_t + b_p,
\end{equation}
where $W_p$ and $b_p$ denote the weight and bias, respectively. Here, $z_t$ represents the predicted counts of locust attacks that will occur in the $(t+1)^{th}$ month for region [x, y].

\section{Experiments}

\subsection{Data set and features}
The input $x_t$ at each step $t$ is a 15-dimensional vector comprising of features described in \ref{DataSource}. For a region, we iteratively provide the model with features of 12 months and predict the number of locust attacks in the $13^{th}$ month. Further, to account for nearby region correlations, at any position (x, y), we sum the number of swarms within a kernel of size $K\times K$. We perform this aggregation only for the locust type ``swarm'' because they are massive, comprise of large number of locusts, and move very quickly. The splitting of the data set is done as descried in Section \ref{DataSource}. In Table~\ref{table:dataStats}, we report the data set statistics. We observe that the number of entries having at least one swarm attack is quite small. Further, scenarios having at least one swarm attack have mean and variance of their counts \textbf{8.52} and \textbf{547.16}, respectively. Data scarcity combined with high variance of output makes the task of training efficient models for this task very challenging.
 
\begin{table}
\caption{Data set statistics}
\begin{center}
\begin{tabular}{l||c|c|c|c}
\label{table:dataStats}
\textbf{Type} & \textbf{Entries} & \textbf{Entries} & \textbf{Start Date} & \textbf{End Date}\\
& (Total) & (\# Swarms > 0) & (yyyy-mm-dd) & (yyyy-mm-dd) \\
\hline
Training & 24092 & 3410 & 1985-01-01 & 2017-05-31\\
Validation & 1830 & 67 & 2017-06-01 & 2019-06-30\\
Test & 3574 & 927 & 2019-07-01 & 2021-07-31\\
\end{tabular}
\end{center}
\end{table}

\subsection{Implementation}
In this work, we used Mean Squared Error (MSE) between original and predicted count of locust attacks as the loss function for training. We trained our model parameters with the Adam \cite{kingma2014adam} optimizer. The batchsize, dimension of \gls{lstm} hidden state, and learning rate was kept to be 64, 32, and $10^{-4}$, respectively. The kernel size, K, is assigned a value of 3. Machine with single NVIDIA V100 GPU with 16 GB memory was used for training. It takes around 30 minutes to train the model for 50 epochs. 

\subsection{Results and Discussion}
In this section, we present and discuss the obtained quantitative as well as the qualitative results.
 \begin{figure}
\centering
\includegraphics[scale=0.6]{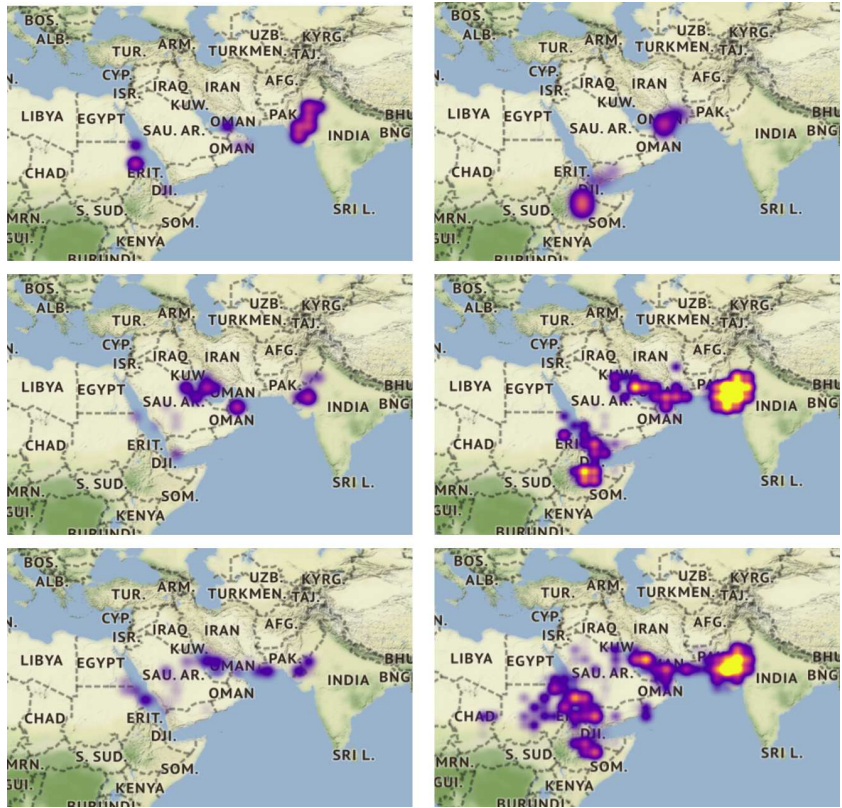}
\caption{Heatmap for comparison between the model's prediction and the ground truth. Here, we present the result for two months, April'20 (left) and July'20 (right). The top row shows the actual observation in the previous month ($t$ - 1). The middle row is the ground truth for $t^{th}$ month, and bottom row denotes the prediction made of the $t^{th}$ month. Yellow, red and purple denote regions with high, medium and low density, respectively.}
\label{fig:heatmap}
\end{figure}
\subsubsection{Quantitative Results}
\label{Quantitative_results}
We evaluate the performance of our model on the standard metrics precision and recall. As the data set is skewed, we measure the macro-average variants of precision and recall. We want to measure how good our model is at predicting the occurrence of a locust attack in a region. Thus, we convert the outputs and ground truth to binary labels. We assign a label of 1 to the predicted output if it is greater than 0.5. Similarly, for ground truth, we take all the entries with at least 1 swarm attack as a positive signal. The obtained macro average recall is \textbf{81\%}. Having a high recall is important as it measures how good we are at predicting a locust attack in advance. However, at the same time, we do not want the model to predict a lot of false alarms. This makes precision an equally important measure. The macro averaged precision score for our model is \textbf{60\%}.

\subsubsection{Qualitative Results} Figure \ref{fig:heatmap} further demonstrates the performance of our model at predicting the count of locust attacks that occur in a particular region after observing the previous 12 months. Firstly, we assess our model based on its ability to predict the extent of swarm infestation. To achieve this purpose, we define a notion of \emph{density} as the number of attacks we observe within a region. Density is an important factor in our assessment since it represents the extent of damage caused by the swarm of locusts in each region. A larger number of swarms in a region indicates more locusts and thus more damage to the area. We define the three density levels low, medium and high in accordance with the number of attacks in each region. In particular, we divide the test entries with at least 1 attack into 3 bins. We label the entries with only 1 attack, between 2 and 4 attacks, and more than 4 attacks as low, medium, and high density bins, respectively. The recall for the bins are 73\%, 75\%, and 86\%, respectively. The quantitative results show that the model has higher recall for high density, as supported by the qualitative results presented in figure \ref{fig:heatmap} as well. 

It is equally important to predict the attacked locations with high precision. In this regard, figure \ref{fig:heatmap} shows a high precision in our prediction. Our results demonstrate that the model is able to find the infested regions as well as the count of locust attacks with good accuracy. Finally, the reason for picking \gls{rnn} for this study is to allow the network to learn time series dependencies. To observe this, we also show the heatmap of attack intensity for the previous month along with the ground truth and prediction for current month. In the second column, we can clearly see that the model is able to capture the movement of locusts from Oman in June'20 to west India in July'21.

\section{Conclusion}
We successfully apply a simple \gls{lstm} model and predict the location of swarms one month in the future based on past observations across the affected regions and obtain a precision and recall of 60\% and 81\%, respectively. As temperature, rainfall and soil moisture are known to influence the swarm movement, future work could be focused on adding information from satellite images which can help include more detailed information regarding humidity and the climate in general. With this additional information, it will be possible to predict the location of the swarm farther in the future and allow the local municipalities to take preemptive measures to reduce the impact of the oncoming swarms and organize early new supply chains to compensate for the damaged crops. The observed data taken on the fields could also be combined to train the model that attempt to detect swarms from satellite images, giving more complete data to analyze as the observation is very sparse and might not include all swarms.

\section*{Acknowledgements}
The authors would like to thank Sirisha Reddy and Fereshteh Shakeri for their valuable feedback and discussions. The authors are also grateful to CIFAR for funding and Compute Canada for computing resources.

\end{document}